# mmJoints: Expanding Joint Representations Beyond $(x, y, z)$ in mmWave-Based 3D Pose Estimation

Zhenyu Wang
University of North Carolina at Chapel Hill
Chapel Hill, USA
zywang@cs.unc.edu

Mahathir Monjur
University of North Carolina at Chapel Hill
Chapel Hill, USA
mahathir@cs.unc.edu

Shahriar Nirjon
University of North Carolina at Chapel Hill
Chapel Hill, USA
nirjon@cs.unc.edu

## ABSTRACT

In mmWave-based pose estimation, sparse signals and weak reflections often cause models to infer body joints from statistical priors rather than sensor data. While prior knowledge helps in learning meaningful representations, over-reliance on it degrades performance in downstream tasks like gesture and activity recognition. In this paper, we introduce mmJoints, a framework that augments a pre-trained, black-box mmWave-based 3D pose estimator's output with additional joint descriptors. Rather than mitigating bias, mmJoints makes it explicit by estimating the likelihood of a joint being sensed and the reliability of its predicted location. These descriptors enhance interpretability and improve downstream task accuracy. Through extensive evaluations using over 115,000 signal frames across 13 pose estimation settings, we show that mmJoints estimates descriptors with an error rate below 4.2%. mmJoints also improves joint position accuracy by up to 12.5% and boosts activity recognition by up to 16% over state-of-the-art methods.

## 1 INTRODUCTION

**mmWave-based 3D human pose estimation suffers from over-reliance on priors bias.** Pose estimation, which estimates 3D body joint coordinates, is a key sensing task that underpins numerous downstream applications, including gesture recognition [36], gait analysis [2], patient monitoring [45], and posture tracking [28], making its accuracy and reliability critical to prevent error propagation. However, mmWave-based pose estimation models [4, 10, 54] exhibit a unique bias problem: *they often rely more on prior knowledge of human body structure and common poses than on actual sensor signals, particularly when radar reflections are sparse.* Unlike vision-based systems that capture rich spatial details, mmWave radars often receive incomplete or noisy reflections [26, 34, 46, 49], leading models to compensate by learning statistical priors instead of interpreting sensor data. Some models even learn to ignore radar signals because prior knowledge statistically results in better overall pose estimation accuracy.

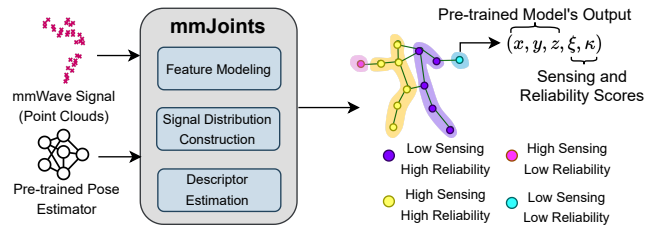

**Figure 1: mmJoints generates descriptors for the estimated joints of pre-trained, black-box models, enhancing the accuracy of downstream tasks.**

**Example.** If lower-body reflections are sporadically missing due to factors such as the positioning of the radar or occlusions in the environment, the model tends to ignore these signals and defaults to memorized lower-body coordinates [4, 10], yielding decent accuracy but lacking sensor-based verification. While this strategy can produce reasonable estimates, it introduces risks in applications requiring precise tracking of task-sensitive joints [58], where misplaced confidence in statistical priors can result in suboptimal or misleading outputs for downstream models [35, 42, 50, 51].

**Priors are vital for learning, but over-reliance skews results.** While model bias is a well-known phenomenon with various mitigation strategies [4, 10, 52–54], it remains an inherent and often necessary component of mmWave sensing. mmWave radars capture highly inconsistent and unreliable reflections from the human body [1, 3, 48], making it practically impossible to position the radar in a way that ensures adequate signal reception from all body parts, even with signal augmentation techniques in software and hardware [52, 60]. A model that depends solely on mmWave signals without incorporating prior knowledge would struggle to learn the concept of human pose, as it lacks dense signals from all joints. Thus, *prior bias is crucial for learning meaningful representations, but over-reliance is detrimental, often causing models to ignore weak or inconsistent radar signals, leading to inaccurate pose estimates that propagate errors to downstream applications.*

**A new proposal—let's make bias explicit.** Instead of attempting to surgically mitigate bias—a task that is both computationally and theoretically challenging due to the stochastic nature of mmWave reflections—we propose a fundamentally different approach: *making bias explicit in the predicted pose*. Specifically, we introduce two additional descriptors for each joint, the *sensing score* ($\xi$) and the *reliability score* ($\kappa$), where $\xi$ represents the extent to which the joint's position is informed by radar signals, and $\kappa$ quantifies the reliability of the estimation. This explicit representation allows downstream tasks to account for signal uncertainty, leading to more accurate activity recognition and improved pose estimation. By leveraging the model's inherent behavior rather than suppressing bias, our approach enables a more robust and interpretable utilization of mmWave-based pose estimation, ultimately enhancing the reliability of human-centric sensing applications.

**Estimating $\xi$ and $\kappa$ is non-trivial, requiring non-trivial, inverse process.** Directly computing $\xi$ and $\kappa$ at inference time is not possible, as by definition, they require ground truth joint locations, which are unavailable once the system is deployed. Standard machine learning methods cannot predict these scores solely from a *(pre-trained model, input signal)* pair, necessitating a more sophisticated approach. To address this, we introduce a *surrogate pose*—a refined pose that best aligns with the input signal and can be estimated using only the available information: i.e., the input signal and the pre-trained model's output. We achieve this through an inverse modeling approach that learns a mapping: ***pose → expected mmWave signal distribution*** to predict the expected sensor data for a given pose. By comparing the observed signal with the expected signal distribution, the discrepancy guides pose refinement in the latent space. This unconventional but effective formulation iteratively adjusts the pose, starting from the pre-trained model's output, to obtain the surrogate pose. As a proxy for the missing ground truth, the surrogate pose enables the inference of sensing and reliability scores.

**We present—"mmJoints".** mmJoints augments the output of a pre-trained pose estimator with a *sensing score* $\xi$ and a *reliability score* $\kappa$, as shown in Figure 1. This enriched 5D joint representation ($x, y, z, \xi, \kappa$) significantly improves the accuracy of downstream tasks such as gesture and activity recognition compared to using the output pose of the pre-trained model without mmJoints-enabled augmentation. mmJoints does not require access to the internal parameters of the pre-trained model. Instead, it treats the pose estimator as a black box. This design allows the pre-trained pose estimator to be hosted locally or accessed through an API. mmJoints has minimal overhead, requiring only 23 ms per inference, with a model size of 75 MB, making it suitable for deployment on edge devices. Notably, mmJoints is not a pose estimator itself. However, when pose estimation is framed as a downstream task, mmJoints can also improve its accuracy.

**Comprehensive evaluation of mmJoints across models, datasets, and real-world scenarios.** We extensively evaluate mmJoints across multiple datasets and pre-trained models, including four state-of-the-art models—MARS [4], mmBody [10], mmMesh [54], and SynMotion [59] and several hand-crafted derivations. Our evaluation uses two public datasets [4, 10] and one self-collected dataset, comprising over 115,000 signal frames across 13 pose estimation settings. Additionally, we conduct real-world experiments with seven pose estimators, 10 activities, and four environments, focusing on two downstream tasks: improving pose estimation and activity recognition. mmJoints runs in real-time on a standard GPU (GTX 1080) with an execution time under 25 ms. It produces accurate descriptors with an error rate below 4.2%, achieves joint position improvements of up to 12.5% over state-of-the-art methods in pose estimation, and boosts activity recognition accuracy by up to 16%.

## 2 MOTIVATION

### 2.1 Limitation of Accuracy Metric

Pose estimation involves predicting the 3D coordinates of 17–22 body joints, such as the head, shoulders , and ankles. Accuracy is measured by joint estimation error—the distance between estimated and ground truth coordinates. A joint is considered *correctly* predicted if this error falls below a threshold; otherwise, it is *incorrect*. However, this metric does not distinguish whether the prediction is based on actual radar signals or inferred from the model's learned priors, reducing insight into the model's decision-making process.

As shown in Figure 2 (left), joints may be predicted correctly despite insufficient radar signals or incorrectly despite receiving sufficient signals. For example, if the radar frequently misses reflections from the lower body, the model may learn to ignore signals from the knees and ankles, inferring their typical positions instead.

### 2.2 Sensing and Reliability Scores

We define two descriptors—the *sensing score* $\xi$ and the *reliability score* $\kappa$—to characterize each joint's radar signal quality and the pre-trained model's confidence in mmWave-based pose estimation. For a given input (mmWave signal), $\xi$ and $\kappa$ depend on both the pose estimated by the pre-trained model and the corresponding ground truth pose.

**Sensing Score ($\xi$).** The sensing score is a measure of radar signal intensity reflected by a joint. Formally, for each joint $j$, we quantify its signal strength as $\psi_j = \sum_{n=1}^{N} \frac{I_n}{d_{j,n}^2}$, where $N$ is the total number of points in the mmWave point cloud, $I_n$

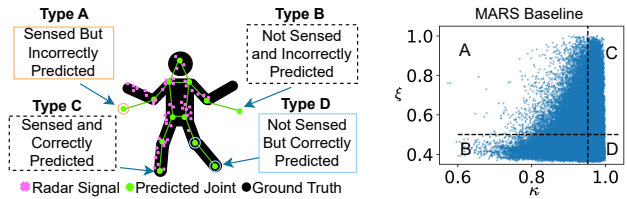

Figure 2: Joint types and their distribution for MARS model.

| Model | Dataset | Correct | | Incorrect | |
|---|---|---|---|---|---|
| | | Not Sensed | Sensed | Not Sensed | Sensed |
| MARS (CNN-depth 6) | MARS | 55,884 | 30,110 | 45,649 | 20,053 |
| MARS (CNN-depth 4) | MARS | 48,657 | 24,118 | 52,876 | 26,045 |
| MARS (CNN-depth 6) | mmBody | 114,867 | 17,338 | 34,145 | 8,462 |
| mmBody (P4T-2 frames) | MARS | 75,837 | 39,532 | 25,111 | 10,437 |
| mmBody (P4T-5 frames) | MARS | 86,588 | 43,177 | 12,805 | 6,010 |
| mmBody (P4T-2 frames) | mmBody | 117,449 | 16,505 | 31,482 | 9,288 |
| mmBody (P4T-5 frames) | mmBody | 118,169 | 17,046 | 30,523 | 8,722 |
| mmMesh (3 frames) | MARS | 76,249 | 35,761 | 25,295 | 14,391 |
| mmMesh (10 frames) | MARS | 90,552 | 43,317 | 11,000 | 6,827 |
| mmMesh (3 frames) | mmBody | 121,194 | 17,395 | 27,818 | 8,405 |
| mmMesh (10 frames) | mmBody | 119,432 | 16,818 | 29,580 | 8,982 |

Table 1: Distribution of joint types across different model variants and datasets.

| Model Input | Accuracy |
|---|---|
| Joint | 75.26% |
| Joint + Descriptor | 86.41% |

(a) Describing joints improves accuracy.

| Joint Type | Correct | | Incorrect | |
|---|---|---|---|---|
| | Not Sensed | Sensed | Not Sensed | Sensed |
| #Joint | 1,078 | 31 | 1,048 | 47 |

(b) A closer look at test cases where joint descriptors help correct mispredictions.

Table 2: Effect of pose characterization on activity recognizer.

is the intensity of the reflected signal at point $n$, and $1/d_{j,n}^2$ denotes the normalized inverse squared distance between joint $j$ and point $n$. A joint is categorized as *sensed* if its signal strength exceeds a predefined threshold.

We define the sensing score $\xi_j$ as:

$$\xi_j = \text{sigmoid}(\psi_j - \bar{\psi}) \tag{1}$$

Here, the sigmoid function bounds and shifts the signal strength relative to the average signal strength $\bar{\psi}$, computed from the training data. Given that signal intensity varies with different mmWave device configurations, we use this dataset-specific average to set a robust reference point. This ensures sufficient variability in the sensing score and prevents saturation across diverse signal conditions.

**Reliability Score ($\kappa$).** The reliability score quantifies the model's confidence in its joint position predictions. Formally, for each joint $j$, we define the reliability score $\kappa_j$ based on the distance $D_j$ between the predicted and ground truth joint positions. To enable practical computation at runtime, we introduce an intermediate mapping:

$$f(D_j) = 1 - \text{sigmoid}\left(D_j - \frac{1}{2}\bar{D}_0\right) \tag{2}$$

where $\frac{1}{2}\bar{D}_0$ is half the average torso length derived from training data. Inspired by the widely adopted PCK@0.5 metric definition [6], which uses body part scales for anatomical measurement, we similarly use half the average torso diameter as a normalization threshold. This ensures that the reliability score remains interpretable and comparable across subjects with varying body sizes. The sigmoid function provides a bounded and stable mapping from the unbounded distance space. Finally, we normalize this mapping as:

$$\kappa_j = \frac{f(D_j)}{f(0)} \tag{3}$$

so that $\kappa_j = 1$ when the predicted joint position perfectly matches the ground truth ($D_j = 0$). This formulation ensures interpretability and variability across different prediction scenarios.

## 2.3 Empirical Study on Model Biases

We conduct an empirical study to quantify biases in state-of-the-art mmWave-based pose estimation models. Ideally, accurate predictions occur when both sensing and reliability scores are high, and inaccurate when both are low. However, we find substantial deviations: many accurate predictions have low sensing scores, indicating reliance on learned priors rather than signal quality, while numerous inaccurate predictions occur despite high scores, suggesting underutilization of available signals. We empirically quantify these deviations using multiple open-source models and datasets.

**Models and Datasets.** We study three state-of-the-art models: MARS [4], mmBody [10], and mmMesh [54] on their publicly available datasets [4, 10]. For fairness and consistency, we train the model on the training dataset, select the best-performing one based on the validation dataset, and then evaluate it on the test dataset. Initial frames are skipped from input sequence to match the mmBody model. For the mmBody dataset, we use the normalized square of signal amplitude as the reflected intensity.

**Distribution of Joint Types.** In Figure 2 (right), joints are plotted based on their sensing and reliability scores. We choose the thresholds for $\xi$ and $\kappa$ at $\psi_j = \bar{\psi}$ and $D_j = \frac{1}{2}\bar{D}_0$, respectively, corresponding to the sigmoid function's midpoint. We observe that 36.84% are correctly predicted despite low sensing scores (region D), while 13.22% are incorrectly predicted despite having high sensing scores (region A).

Table 1 summarizes the joint distributions across three models and their variants on multiple datasets, highlighting that a substantial number of correctly estimated joints lack radar signal support. On MARS, the average number of *not sensed but correctly* (*sensed but incorrectly*) predicted joints across various baselines is 72,295 (13,961), accounting for 66.76% (32.66%) of all correctly (incorrectly) predicted joints. A similar trend is observed in the mmBody dataset. Additionally, the distribution of correctly and incorrectly predicted joints varies with model accuracy, influenced by signal sufficiency.

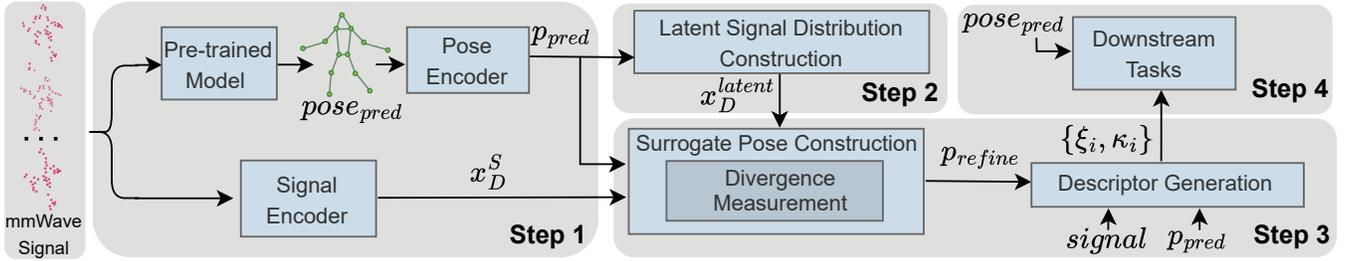

Figure 3: mmJoints consists of four steps. After pose and signal modeling (Step 1), we construct the inverse function (Step 2) for surrogate pose construction and descriptor generation (Step 3). The estimated descriptors enhance downstream tasks (Step 4).

## 2.4 Does $(\xi, \kappa)$ Help Improve Downstream Task's Accuracy?

We develop a 10-class activity recognizer using 3D coordinates of 19 body joints as input, with poses estimated by the pre-trained MARS [4] model. The activities, sourced from the MARS dataset, include both upper and lower body actions. The baseline recognizer achieves 75.26% accuracy. Enhancing it with joint descriptors $\xi$ and $\kappa$ as additional inputs (modifying only the input layer for fairness) improves accuracy by 11.15%. To explain this boost, we analyze test cases where the baseline model fails but the enhanced classifier succeeds. Table 2 (b) shows that unsensed joints constitute a significant portion of both correctly and incorrectly predicted joints.

This suggests that the enhanced classifier leverages sensing and reliability scores to accurately identify activities despite sparse input signals. In contrast, the baseline model, relying on statistical priors when signals are insufficient, produces poses with many correct or incorrect joints that are not sensed. $\xi$ and $\kappa$ guide understanding of learnable patterns in the behavior of the model, providing valuable information to improve the accuracy of the downstream application.

## 3 OVERVIEW AND CHALLENGES

### 3.1 System Overview

mmJoints takes as input a pre-trained pose estimation model and an mmWave signal (e.g., point cloud) and enhances the model's output by adding two descriptors for each joint: a *sensing score* and a *reliability score*. Combined with the pre-trained model's 3D coordinates, these descriptors form an enriched *5D joint representation* $(x, y, z, \xi, \kappa)$, providing insight into each joint's estimation process.

Since ground truth joint locations are required for direct computation of $\xi$ and $\kappa$ but are unavailable at inference time, mmJoints estimates a *surrogate pose*—a refined pose in a learned latent space that best fits the given signal—using the pre-trained model's output pose as a seed. This refined *surrogate pose* serves as a proxy for the ground truth, enabling mmJoints to estimate the sensing and reliability scores.

The key technique is to learn an *inverse function* that maps poses to their expected signals, predicting *what the sensor data should look like for a given pose*. By comparing the observed signal to this expected signal distribution, mmJoints computes the difference and adjusts the pose in the latent space, refining it toward a more accurate estimate. Empirically, a single iteration of this correction is sufficient. All computations occur in the latent space, ensuring robustness to noise and uncertainty.

**Processing Steps:** mmJoints consists of four key steps as shown in Figure 3 which are briefly described next:

- **Step 1:** (a) A fixed number $K$ of input signal instances (mmWave point clouds) are processed to extract features and fitted to a Gaussian distribution, yielding the distribution-based signal feature $x_D^S$. (b) The estimated pose $pose_{pred}$ from the pre-trained model undergoes feature extraction to obtain the pose feature $p_{pred}$.

- **Step 2:** A latent signal distribution $x_D^{latent}$ is generated from the pose feature $p_{pred}$. Further details and motivations are in Section 5.

- **Step 3:** (a) An approximation of the Jeffrey divergence [24] between $x_D^S$ and $x_D^{latent}$ is computed to quantify their relative displacement. (b) The divergence, input signal feature, and estimated pose feature are used to generate the refined latent pose $p_{refine}$. (c) $p_{pred}$, $p_{refine}$, and the current input signal instance are fed into a descriptor generation block to estimate the sensing and reliability scores $(\xi, \kappa)$ for each joint of the black-box model's output pose $pose_{pred}$.

- **Step 4:** The descriptors $\{\xi_i, \kappa_i\}$ are appended to $pose_{pred}$, providing an enriched input for downstream tasks such as activity recognition and high-precision pose estimation.

### 3.2 Technical Challenges

Three major challenges are solved by mmJoints:

- **Feature Modeling.** mmWave signals (e.g., point clouds) and pose joint coordinates in raw space are challenging to parameterize and utilize directly due to their unstructured and complex nature. To address this, we extract essential features from the input signal and pose using specific strategies and process them in a continuous latent space, enabling the model to learn a mathematically tractable representation of the data.

- **Signal Distribution Construction.** mmJoints leverages signal distribution information, which correlates more strongly with pose than individual signal instances (details in Section 5). Constructing this distribution from an arbitrary pose is challenging, as it must remain structured and tractable for model learning and measurements (e.g., computing divergence). To address this, we propose a novel approach that constructs a closed-form distribution in latent space and introduces a methodology for representing arbitrary pose types using suitable learning principles.

- **Refined Pose Representation.** Utilizing the constructed signal distribution to generate a precise pose representation is challenging, as the measurement must remain tractable with reasonable approximations. In mmJoints, we exploit the properties of the modeled signal and pose features to compute the *displacement* between the input signal and the signal distribution of the pre-trained model's output. This approach produces a latent pose that is robust to variations in black-box model performance and serves as an input for descriptor estimation.

# 4 SIGNAL AND POSE MODELING
## 4.1 Pose Modeling

**Rationale.** Our goal is to model a broad and diverse range of human poses. Representing each pose as a list of 3D coordinates $(x, y, z)$ is neither scalable nor analytically tractable. Instead, we seek a parametric representation in which any pose can be expressed as a linear combination of a small set of *basis* poses. This enables efficient interpolation, generalization, and manipulation of poses.

**Processing Steps.** Pose modeling consists of three steps:

- **Step 1:** We cluster all pose instances from the training dataset using a Gaussian Mixture Model (GMM), selecting the number of clusters $C$ ($C \geq D_p$) based on the Bayesian Information Criterion (BIC) [47], where $D_p$ is the dimension of the latent feature space. Each cluster represents a distinct pose category. We use $d_i \times |\{pose_i\}|$ as the metric to select the top-$D_p$ clusters for basis construction, where $|\{pose_i\}|$ is the number of the poses in the $i^{th}$ cluster, and $d_i$ is the Euclidean distance between the centroid of $i^{th}$ cluster and the mean pose.

- **Step 2:** To effectively represent raw poses while maintaining continuity and smoothness, we employ a Variational Auto-Encoder (VAE) [21, 27]. The output features follow a Gaussian distribution, characterized by the mean pose feature $\mu_p$ and standard deviation $\sigma_p$. During training, we use the objective function:

$$\mathcal{L}_{P_{step2}} = \mathcal{L}_{REC} + \lambda_{step2}\mathcal{L}_{KL} \quad (4)$$

, where $\mathcal{L}_{REC}$ is the reconstruction loss implemented with Mean Squared Error (MSE) to ensure the latent features capture the input's key characteristics; $\mathcal{L}_{KL}$ regularizes the latent feature space by penalizing $\mathcal{N}(\mu_p, diag(\sigma_p)^2)$ towards the multivariate normal distribution[1].

- **Step 3:** The $D_p$ pose categories are designed to be mutually orthogonal in the latent space to generate a valid pose basis (details in Section 5), which is fine-tuned using the Orthogonal Projection Loss (OPL) [43]:

$$\mathcal{L}_{P_{step3}} = \mathcal{L}_{OPL-d} + \lambda_{step3}\mathcal{L}_{OPL-s} \quad (5)$$

, where it emphasizes inter-class orthogonality while ensuring the quality of latent feature clustering.

## 4.2 Signal Modeling

**Rationale.** In mmWave-based pose estimation, the input signal is typically a multi-dimensional, unordered, and sparse point cloud [4, 10, 54]. Standard encoder-decoder models [27] struggle with such sparse inputs, resulting in unstable reconstructions. To address this, we transform the signal into a structured feature space, where signals corresponding to similar poses are mapped closer together, while those representing different poses are mapped farther apart.

**Processing Steps.** We use a Point-4D-Transformer-based feature extractor [16], designed for point cloud inputs. It generates the distribution feature $x_D^S$, which consists of the mean $\mu_S$ and standard deviation $\sigma_S$, forming a Gaussian distribution to represent consecutive $K$ signal instances. The objective function for signal modeling process is:

$$\mathcal{L}_S = \lambda_0 \mathcal{L}_{triplet} + \lambda_1 \mathcal{L}_{CE} + \lambda_2 \mathcal{L}_{KL} \quad (6)$$

, where $\mathcal{L}_{triplet}$ is the Triplet Loss [7] for contrastive learning over the latent feature space to capture the correct instance-wise correlation; $\mathcal{L}_{CE}$ is the cross-entropy loss function that ensures cluster-wise correlation labeled by pose clusters; $\mathcal{L}_{KL}$ penalizes the difference between two multivariate distributions: $\mathcal{N}(\mu_S, diag(\sigma_S)^2)$, $\mathcal{N}(0, I)$, regularizing the latent space for $x_D^S$; $\lambda_0$, $\lambda_1$ and $\lambda_2$ indicate the contribution of these three loss terms.

## 4.3 Signal Distribution Modeling

**Rationale.** Recall that in our inverse modeling approach, we map poses to signal distributions. The justification for this approach is provided in Section 5.1. To represent the distribution, we extract feature representations of signal instances and model them using a Gaussian Mixture Model.

---
[1]We follow MATLAB convention for $diag(\cdot)$: applied to a matrix, it extracts the diagonal into a vector; applied to a vector, it forms a diagonal matrix.

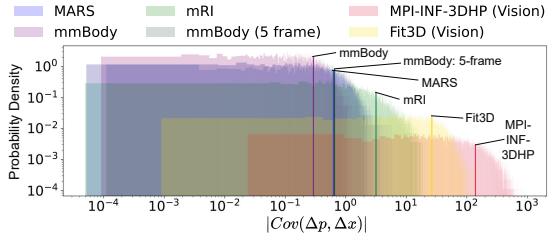

Figure 4: Covariance analysis: signal vs pose.

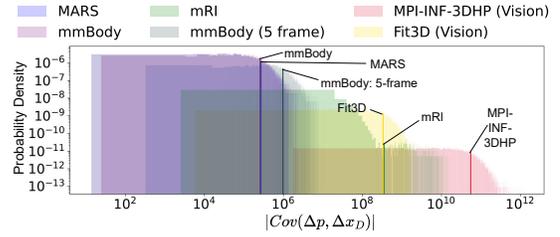

Figure 5: Covariance analysis: signal distribution vs pose.

**Processing Steps.** Signal distribution modeling consists of two key steps as below:

- **Step 1:** For each pose instance $pose_i$, we select the top $M$ closest poses, up to a maximum of $M_{max}$, by measuring their distances in the latent feature space using the $l_2$-norm between their respective $\mu_p$. These distances are limited to $||\sigma_{p_i}/2||$. By collecting the corresponding $M$ signal inputs and their associated latent features, we obtain $M$ normal distribution components:

$$\{\mathcal{N}(\mu_{S_i}, diag(\sigma_{S_i})^2) \mid i = 1, 2, ..., M\} \quad (7)$$

- **Step 2:** To formalize the distribution representation for effective training, rather than using a nonstandard set form, we construct a GMM with $G$ components, using samples randomly drawn from distribution components from Equation (7) — $\mathcal{N}(\mu_{S_i}, diag(\sigma_{S_i})^2)$. The value of $G$ selected based on the BIC indicator applied to a randomly sub-sampled dataset, with an upper bound to prevent overfitting. This results in a closed-form distribution:

$$\mathcal{P}(\mathbf{x}) = \sum_{i=1}^{G} \phi_i \mathcal{N}(\mathbf{x} \mid \mu_i, \Sigma_i), \ s.t. \sum_{i=1}^{G} \phi_i = 1 \quad (8)$$

with respect to the latent signal feature **x**. Similar to the proposition in [27], we use diagonal covariance for each component to maintain an efficient modeling strategy.

## 5 LATENT SIGNAL DISTRIBUTION CONSTRUCTION

### 5.1 Why Signal Distribution?

**Rationale.** Signal distribution exhibits a stronger relationship to pose when examined via covariance analysis. This justifies our inverse modeling approach, where we map pose → signal distribution.

**Signal Instance vs Pose.** We systematically study the pose estimation problem via covariance analysis [15] where we quantify the correlation strength of the changes to the signal, $\Delta x$ and the corresponding changes to the ground truth pose, $\Delta p$. A larger absolute covariance between $\Delta p$ and $\Delta x$ indicates a stronger dependency between $x$ and $p$, which is relatively easier to model.

Figure 4 compares the distribution of $|Cov(\Delta p, \Delta x)|$ for four mmWave-based and two computer vision-based pose estimation baselines [3, 4, 10, 18, 38]. The vertical lines represent the mean. For consistency across data domains, $\Delta x$ is computed using the $l_2$-norm of input signal vectors before they are fed to the pose estimator; $\Delta p$ denotes the Euclidean distance between two poses. We observe that mmWave baselines are scattered around the left side of the plot, while the computer vision baselines coincide on the right.

This stark difference between vision and mmWave baselines can be attributed to the nature of signal reflections. mmWave, when reflected off the human body, are sparse, sporadic, and inconsistent. In contrast, light uniformly illuminates the body, resulting in dense, continuous, and consistent reflections. As a result, mmWave radars face significantly greater challenges in achieving the same level of performance as cameras in accomplishing human sensing tasks.

**Signal Distribution vs Pose.** We can lower the modeling complexity by reformulating the problem. We replace signal instance, $x$ with its distribution, $x_D$ and study $|Cov(\Delta x_D, \Delta p)|$. We redo the covariance analysis for this new formulation to observe the dependency of changes in the ground truth pose, $\Delta p$ and the corresponding signal distribution change, $\Delta x_D$. To obtain the distribution $x_D$ from a collection of signal instances given a pose $p$, we first apply PCA to reduce the number of correlated features in the raw signal space and then fit this to a GMM. Symmetric KL-divergence [24] is used to compute $\Delta x_D$.

Figure 5 shows that the mean $|Cov(\Delta x_D, \Delta p)|$ for both vision and mmWave baselines are overlapping at the right end with much larger variance values. *This indicates that the signal distribution and pose maintain a stronger relationship compared to the association between pose and signal instances.* Despite their strong relationship, at runtime, a standalone radar system cannot sense the signal distribution directly. While some works attempt to approximate this by accumulating consecutive frames [3, 5], these frames are biased by their fixed location and do not produce high-quality signal distributions. To address this limitation, we introduce latent signal distribution in mmJoints, with the details of its construction provided in Section 5.2.

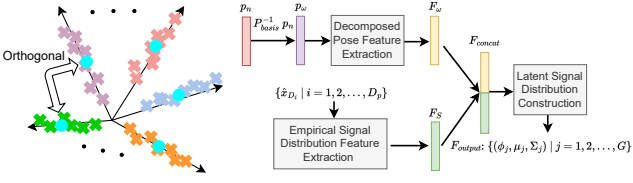

Figure 6: Pose basis construction.

Figure 7: Illustration of latent signal distribution construction.

## 5.2 Latent Signal Distribution Construction

**Rationale.** Our goal is to construct an analytically tractable, closed-form representation of the signal distribution for any arbitrary input pose. To achieve this, we first define a pose basis consisting of a fixed set of orthogonal poses, allowing the input pose to be expressed in this basis. A model is then trained to map the input pose to its corresponding expected signal distribution. The empirical signal distribution data corresponding to the pose basis provide a stable reference that aids model convergence.

**Processing Steps.** The four major steps in this module are shown below:

- **Step 1:** The basis poses are obtained by modeling the pose instances from the $D_p$ clusters (as described in Section 4.1), representing the pose space as

$$\{(F_{basis_i}, \hat{x}_{D_i}) \mid i = 1, 2, ..., D_p\} \quad (9)$$

, where $F_{basis_i}$ represents the average of $\{\mu_p\}$ of the $i^{th}$ pose cluster, such that all the $\{F_{basis_i}\}$ are orthogonal to each other based on the distributive property of dot product operation. $\hat{x}_{D_i}$ is the input signal distribution corresponding to $F_{basis_i}$ to provide empirical knowledge, which is represented as $\{(\phi^i_{basis_j}, \mu^i_{basis_j}, \Sigma^i_{basis_j}) \mid j = 1, 2, ..., G\}$. In Figure 6, the cyan features represent the mean of each pose cluster's features. Thus, we define the pose basis as

$$P_{basis} = [F_{basis_1}, F_{basis_2}, ..., F_{basis_{D_p}}] \quad (10)$$

, where $P_{basis} \in \mathbb{R}^{D_P \times D_P}$ is an invertible matrix.

- **Step 2:** In Figure 7, we show the detailed flow of the latent signal distribution construction module. For an arbitrary pose $pose_n$, it is characterized as $p_\omega$ by decomposing its latent feature $p_n$,

$$p_\omega = P_{basis}^{-1} p_n \quad (11)$$

, from which the fine-tuned feature $F_\omega$ is then extracted.

- **Step 3:** The $\{\hat{x}_{D_i}\}$ is used as empirical data to aid in mitigating the difficulty of signal distribution construction and facilitates convergence. We use the SetTransformer [30] as the feature extractor to handle the unstructured data format. The input is in the shape of $D_P \times D_S \times G \times D_C$, where $D_S$ is the latent space dimension of the input signal and $D_C$ is the dimension of the concatenation of $\phi_{basis_j}, \mu_{basis_j}$ and $diag(\Sigma_{basis_j})$. This fused representation remains accessible during inference. The empirical feature $F_S$ extracted from $\{\hat{x}_{D_i}\}$ is concatenated with $F_\omega$ and fed into the latent signal distribution generator to obtain the representation of the $G$ mixture components. The output, $F_{output}$, is a flattened feature formed by concatenating $\{(\phi_j, \mu_j, \sigma_j^{\circ 2}) \mid j = 1, 2, ..., G\}$, where $(\cdot)^{\circ(\cdot)}$ denotes the Hadamard power.

- **Step 4:** To achieve accurate estimation of $\{(\phi_j, \mu_j, \sigma_j^{\circ 2}) \mid j = 1, 2, ..., G\}$, the precision measurement used as a loss criterion must be permutation-invariant during training. Let $T(\cdot)$ be an assessment function that takes the distribution component $\Gamma_j = (\phi_j, \mu_j, \Sigma_j)$ as input, we have

$$\mathcal{L}_{LAP} = \min \sum_{1 \leq i,j \leq G} T(\Gamma_i, \Gamma_j) a_{i,j} \quad (12)$$

$$s.t. \sum_j a_{i,j} = 1, \sum_i a_{i,j} = 1, \forall\, i, j \in [1, G] \quad (13)$$

, where $a_{i,j} \in \mathbb{Z}$ and $0 \leq a_{i,j} \leq 1$. This optimization problem can be directly solved using the Hungarian algorithm [37] due to its similarity to the Linear Assignment Problem (LAP). Although the algorithm has cubic complexity, the small size of $G$ ensures manageable computation in practice. During training, $T(\cdot)$ is performed as a combination of three loss terms:

$$T(\Gamma_i, \Gamma_j) = \mathcal{L}_{l_1-\phi} + \mathcal{L}_{l_1-\mu} + \mathcal{L}_{l_1-\Sigma} \quad (14)$$

, where $\mathcal{L}_{l_1-\phi}$, $\mathcal{L}_{l_1-\mu}$, and $\mathcal{L}_{l_1-\Sigma}$ represent the $l_1$ loss function applied on $(\phi_i, \phi_j)$, $(\mu_i, \mu_j)$, and $(\Sigma_i, \Sigma_j)$, respectively;

As such, the loss function for training is:

$$\mathcal{L}_{Distribution} = \mathcal{L}_{LAP} + \lambda_{Div} \mathcal{L}_{Div} \quad (15)$$

, where $\mathcal{L}_{Div} = \hat{J}(\mathcal{P}_S, \mathcal{P}_G)$ (right-hand side of Equation (18)) represents the approximate *displacement* between the current signal feature ($\mathcal{P}_S$) and estimated distribution ($\mathcal{P}_G$), which is explained in detail in Section 6.

## 6 DESCRIPTOR ESTIMATION

### 6.1 Divergence Analysis

**Rationale.** A closed-form metric is defined that quantifies the discrepancy between the expected signal distribution and the current signal input. This metric is used in later stages of surrogate pose construction.

**Processing Steps.** The three key steps in this analysis are shown below:

- **Step 1:** To begin with, we consider an input signal feature $x_D^S$ in a Gaussian distribution form with mean $\mu_S$ and standard deviation $\sigma_S$. The displacement between this representation and the latent signal distribution $x_D^{latent}$ is measured using Jeffrey divergence:

$$J(\mathcal{P}_S, \mathcal{P}_G) = KL(\mathcal{P}_G || \mathcal{P}_S) + KL(\mathcal{P}_S || \mathcal{P}_G) \quad (16)$$

, where $KL(\cdot)$ is the KL-divergence [29]; $\mathcal{P}_S$ is the posterior probability from Gaussian distribution $\mathcal{N}(\mu_S, diag(\sigma_S)^2)$; $\mathcal{P}_G$ is the posterior probability from $x_D^{latent}$ in GMM form: $\mathcal{P}_G(\mathbf{x}) = \sum_{i=1}^{G} \phi_i \mathcal{N}(\mathbf{x} \mid \mu_i, \Sigma_i)$ s.t. $\sum_{i=1}^{G} \phi_i = 1$ ($\phi_i \in [0,1]$). Our goal is to leverage information from $J(\mathcal{P}_S, \mathcal{P}_G)$ to generate surrogate pose that better aligns with signal space.

- **Step 2:** However, there is no analytically tractable Jeffrey / KL divergence when GMM is involved. Thus, we utilize the approximation of the Equation (16), which serves as an upper bound with a tractable closed-form expression. For $J(\mathcal{P}_S, \mathcal{P}_G)$, we have

$$\begin{aligned} J(\mathcal{P}_S, \mathcal{P}_G) &= \int \mathcal{P}_G \ln\left(\frac{\mathcal{P}_G}{\mathcal{P}_S}\right) dx + \int \mathcal{P}_S \ln\left(\frac{\mathcal{P}_S}{\mathcal{P}_G}\right) dx \\ &= \int \left(\sum_{i=1}^{G} \phi_i \mathcal{P}_c^i\right) \ln\left(\left(\sum_{i=1}^{G} \phi_i \mathcal{P}_c^i\right)\Big/\mathcal{P}_S\right) dx \\ &+ \int \mathcal{P}_S \ln\left(\mathcal{P}_S \Big/ \left(\sum_{i=1}^{G} \phi_i \mathcal{P}_c^i\right)\right) dx \end{aligned} \quad (17)$$

, where $\mathcal{P}_c^i$ is the probability from $i^{th}$ mixture component of GMM; since $\mathcal{P}_S$ and $\mathcal{P}_c^i$ are from normal distribution, they are always greater than zero.

- **Step 3:** We consider a convex function with respect to x: $f(x) = J(\mathcal{P}_S, x) = x \ln(x/\mathcal{P}_S) + \mathcal{P}_S \ln(\mathcal{P}_S/x)$ ($x > 0$). We have the following derivations based on Jensen's inequality [25] and Equation (17):

$$\begin{aligned} J(\mathcal{P}_S, \mathcal{P}_G) &= \int f\left(\sum_{i=1}^{G} \phi_i \mathcal{P}_c^i\right) dx \le \int \left(\sum_{i=1}^{G} \phi_i f(\mathcal{P}_c^i)\right) dx \\ &= \sum_{i=1}^{G} \phi_i \int f(\mathcal{P}_c^i) dx = \sum_{i=1}^{G} \phi_i J(\mathcal{P}_S, \mathcal{P}_c^i) \end{aligned} \quad (18)$$

Obviously, the approximation term (weighted Jeffrey divergence) on the right-hand side of the inequality is analytically tractable and can be used effectively for surrogate pose construction.

### 6.2 Surrogate Pose Construction

**Rationale.** By analyzing the displacement between the expected signal distribution and the current signal, we refine the pose obtained from the pre-trained model, resulting in a more accurate pose estimate in the latent space. This refined estimate, known as the surrogate pose, serves as a proxy when the ground truth is unavailable during inference.

**Processing Steps:** The two major steps in this module are shown below:

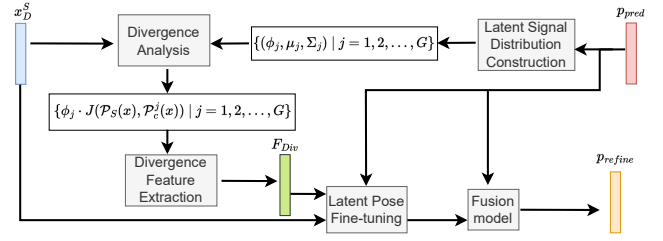

Figure 8: Illustration of surrogate pose construction module.

- **Step 1:** We first define the inputs for the surrogate pose construction module. In addition to the approximated divergence term, we use the features from the pre-trained model's output to facilitate the refinement process. We also use $\mu_S$ from the signal feature $x_D^S$ as an anchor to enhance the distribution representation.

- **Step 2:** The *displacement*—$\{\phi_i J(\mathcal{P}_S, \mathcal{P}_c^i) \mid i = 1, 2, \ldots, G\}$, is preprocessed by a SetTransformer to extract the divergence feature $F_{Div}$. A scaling coefficient is applied to this weighted Jeffrey divergence before being fed into SetTransformer to avoid overflow. After generating the intermediate fine-tuned pose feature, we pass it through a fusion module along with the feature of pre-trained model's output to generate the surrogate pose $p_{refine}$. The corresponding objective function is:

$$\mathcal{L}_{refine} = \mathcal{L}_{huber-final} + \mathcal{R} \quad (19)$$

, where $\mathcal{L}_{huber-final}$ is the Huber Loss [23] between latent feature of ground truth pose and surrogate pose. The regularization term $\mathcal{R} = \lambda_{inter} \mathcal{L}_{huber-inter}$ enhances the optimization by applying constraints to the intermediate fine-tuned pose feature before it is fed into the fusion model, with a strength controlled by $\lambda_{inter}$.

### 6.3 Descriptor Generation

**Rationale.** We estimate the sensing ($\xi$) and reliability ($\kappa$) scores using the input signal data, pre-trained model's output, and the surrogate pose.

**Processing Steps.** We use a two-branch Hourglass-like architecture [41], but with linear layers instead, to estimate both the sensing score and reliability score, leveraging $p_{refine}$, $p_{pred}$, and feature of current signal frame (extracted from a SetTransformer). The optimization process combines sensing and reliability scores, with the training objective function defined as their summed MSE:

$$\mathcal{L}_{desc} = \mathcal{L}_{MSE-\xi} + \mathcal{L}_{MSE-\kappa} \quad (20)$$

These estimated scores, along with the predicted pose from the pre-trained model, are fed into the downstream task to enhance application performance.

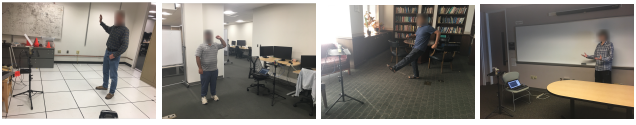

Figure 9: The Self-Collected dataset contains data from a lab area, an office, a library, and a classroom.

| Datasets | Frames for Training | Frames for Validation | Frames for Testing | Joints | Radar |
|---|---|---|---|---|---|
| MARS | 24,066 | 8,033 | 7,984 | 19 | TI IWR1443 |
| mmBody | 27,821 | 4,125 | 7,946 | 22 | Phoenix |
| Self-Collected | 21,864 | 5,476 | 7,706 | 17 | TI AWR1843 |

Table 3: Dataset details.

## 7 EVALUATION

### 7.1 Experimental Setup

**Datasets and Models.** We conduct experiments on three mmWave-based pose estimation datasets: MARS [4], mmBody [10] and Self-Collected datasets. For Self-Collected dataset[2], we collect data from five subjects in four different environments and from two different radar positions using an off-the-shelf commodity AWR1843 mmWave radar [1]. The setup is shown in Figure 9. The dataset contains nine standard activities: left/right/both lateral raises, left/right/both bicep curls, half squats, left/right kicks, and freestyle exercises. The radar captures raw signals, which are converted into point clouds, with each point characterized by $x$, $y$, and $z$ coordinates, as well as Doppler velocity and energy feature, following the preprocessing procedures described in [10, 54]. For the mmBody dataset, we use the first 14 sequences for training, the next two sequences for validation, and the following four sequences for testing. For the Self-Collected dataset, the test dataset includes both unseen environmental and new human subject settings, ensuring thorough assessment of generalizability. In Table 3, we show the details of the datasets. The validation phase follows the setup in [8] to regulate the training process.

We use four state-of-the-art baseline models: MARS (CNN) [4], mmBody (P4Transformer) [10], mmMesh [54], and SynMotion (Attention-based Actual Tracker) [59]. The MARS baseline takes a single frame as input, the mmBody baseline uses five consecutive frames, and the mmMesh and SynMotion baselines use 10 consecutive frames. For SynMotion, we use heatmap [54] as input, which is available in the Self-Collected dataset. We use the corresponding skeletal pose loss function in the training procedure of all baselines.

**Configurations and Metrics.** We use $D_p = 32$, $D_S = 64$, $\bar{D}_0 = 20$, and $D_C = 129$, where the sizes of $\mu_S$ and $\sigma_S$ are both 64. $K$ is determined based on the sampling rate of the radar configuration and corresponds to half-second intervals of the signal. The Adam optimization algorithm is used for training. The evaluation metrics are defined as follows:

- *Weighted Mean Absolute Percentage Error (wMAPE) [40]:* A variation of Mean Absolute Percentage Error (MAPE) [11] offering improved stability for near-zero ground truth.
- *Symmetric Mean Absolute Percentage Error (sMAPE) [11]:* A bounded MAPE variation using symmetric percentage errors to measure prediction accuracy.
- *Mean Per Joint Position Error (MPJPE) [3]:* The mean Euclidean distance between predicted and ground-truth joint positions to evaluate pose estimation accuracy.
- *Mean Absolute Error (MAE) [4]:* The average $l_1$-norm of errors across the $x$, $y$, and $z$ axes for estimated joints.
- *Correct Joint Percentage (PCK@0.5) [6]:* The percentage of joints with prediction errors within 50% of torso diameter.
- *Basis Reconstruction Error:* The representation capability of $P_{basis}$ for input feature $F$, defined as $||F - P_{basis}P_{basis}^T F||_2^2$.
- *Margin Difference:* The representation capability of signal feature $\mu_S$ is evaluated by comparing latent margins to positive signal $\mu_S^{pos}$ (similar pose) and negative signal $\mu_S^{neg}$ (dissimilar pose) representations, formulated as $||\mu_S - \mu_S^{neg}||_2^2 - ||\mu_S - \mu_S^{pos}||_2^2$; higher values indicate better representation.

---
[2]We use stereo cameras and depth camera to construct the ground truth of human pose by following the procedure proposed in [3]

| Datasets | Random Basis Reconstruction Error ($\times 10^3$) | mmJoints Basis Reconstruction Error ($\times 10^3$) | CCS-based Signal Margin Difference | mmJoints Signal Margin Difference |
|---|---|---|---|---|
| MARS | 9.35 | **0.19** | 7.75 | **36.19** |
| mmBody | 16.90 | **2.66** | 13.17 | **22.77** |
| Self-Collected | 5.03 | **0.11** | 7.50 | **33.36** |

Table 4: Pose and signal feature representation evaluation.

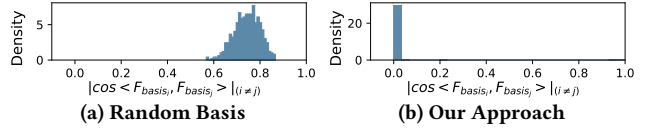

(a) Random Basis | (b) Our Approach

Figure 10: Pose basis orthogonality in MARS dataset.

### 7.2 Component Evaluation

**Feature Modeling.** We evaluate the feature representation capability of the pose and signal encoders. Table 4 shows that the mmJoints basis results in a 49.07× lower reconstruction error than the random basis on the MARS dataset. Figure 10 shows that the basis from our proposed pose modeling is strongly orthogonal, with the cosine values between all possible pairs of basis vectors clustering near zero. We compare our signal modeling method to a baseline using the encoder architecture from [4] but with only coarse cluster-wise structure (CCS) of the input signal, which shows a 4.67× smaller margin difference than our approach. A similar trend across datasets highlights the effectiveness of feature modeling.

**Latent Signal Distribution.** We evaluate the performance of latent signal distribution construction by comparing it with a baseline that uses the same architecture, takes the pose feature as input and optimizes with KL-Divergence between the current signal feature and the target signal distribution, without considering the component-wise inner data structure. We use the state-of-the-art pose estimator, SynMotion, as a pre-trained model and evaluate the accuracy of the estimated distribution components' mean, covariance, and corresponding GMM weights on the Self-Collected dataset. As shown in Table 5, our proposed method consistently achieves better accuracy in all aspects, with improvements of 67.72%, 13.12%, and 39.85% in mean, covariance, and weights, respectively, demonstrating its effectiveness. The impact of the latent signal distribution is discussed in more detail in Section 7.4.

| Configurations | MSE-mean ($\mu \times 10^{-1}$) | MSE-covariance ($\Sigma \times 10^{-3}$) | MSE-weights ($\phi \times 10^{-2}$) |
|---|---|---|---|
| KL Divergence-Based | 14.50 | 5.59 | 2.03 |
| mmJoints | **4.68** | **4.86** | **1.22** |

Table 5: Evaluation of latent signal distribution construction.

| Dataset | Model | Pre-trained MSE ($\times 10^{-2}$) | Refined mmJoints MSE ($\times 10^{-2}$) |
|---|---|---|---|
| MARS | mmBody | 3.81 | **2.50** |
| mmBody | MARS | 22.48 | **15.38** |
| Self-Collected | mmMesh | 6.63 | **6.02** |

Table 6: Evaluation of surrogate pose construction.

| Dataset | Model | $sMAPE_\xi$ (%) | $sMAPE_\kappa$ (%) | $wMAPE_\xi$ (%) | $wMAPE_\kappa$ (%) |
|---|---|---|---|---|---|
| MARS | MARS | 2.23 | 1.81 | 3.37 | 1.77 |
|  | mmBody | 2.31 | 1.52 | 3.56 | 1.49 |
|  | mmMesh | 2.18 | 1.35 | 3.37 | 1.32 |
| mmBody | MARS | 0.86 | 1.70 | 1.28 | 1.66 |
|  | mmBody | 0.94 | 1.93 | 1.39 | 1.89 |
|  | mmMesh | 1.21 | 1.94 | 1.79 | 1.91 |
| Self-Collected | MARS | 2.36 | 2.12 | 3.45 | 2.09 |
|  | mmBody | 2.40 | 3.59 | 3.58 | 3.59 |
|  | mmMesh | 2.30 | 2.60 | 3.41 | 2.59 |
|  | SynMotion | 2.40 | 4.19 | 3.57 | 4.20 |

Table 7: Descriptor estimation evaluation of mmJoints on state-of-the-art baselines.

| Models | Upper Body | | Lower Body | |
|---|---|---|---|---|
|  | MPJPE (cm) | MAE (cm) | MPJPE (cm) | MAE (cm) |
| $M_1$ | 10.54 | 4.98 | 47.06 | 20.56 |
| $M_2$ | 58.60 | 24.77 | 8.67 | 4.05 |
| $M_3$ | 58.69 | 24.80 | 47.60 | 20.98 |

Table 8: Specially-crafted models on Self-Collected dataset.

| Model | $sMAPE_\xi$ (%) | $sMAPE_\kappa$ (%) | $wMAPE_\xi$ (%) | $wMAPE_\kappa$ (%) |
|---|---|---|---|---|
| $M_1$ | 2.48 | 2.16 | 3.42 | 2.13 |
| $M_2$ | 2.60 | 2.62 | 3.46 | 2.57 |
| $M_3$ | 2.74 | 2.61 | 3.48 | 2.59 |

Table 9: Descriptor estimation evaluation of mmJoints on specially-crafted baselines.

**Surrogate Pose Construction.** We evaluate the performance of the surrogate pose construction module by comparing it with a baseline that directly encodes the pre-trained model's output as surrogate pose. As shown in Table 6, for the MARS dataset, where the pre-trained model is mmBody, we observed a 34.30% reduction in error when constructing the surrogate pose. A similar trend is observed across other configurations, demonstrating its efficacy. Section 7.4 details the impact of the surrogate pose in mmJoints

## 7.3 System Evaluation

**mmJoints on State-of-the-Art Models.** We apply mmJoints to state-of-the-art pre-trained models to evaluate the accuracy of the estimated descriptors. Table 7 presents the performance results for estimating the sensing and reliability scores across all model-dataset combinations. For the Self-Collected dataset, mmJoints estimates the sensing score with an average wMAPE of 3.41%—3.58% across the four state-of-the-art baselines. The estimated reliability score also achieves low wMAPE values ranging from 2.09% to 4.20%. A similar trend across different datasets and baselines, measured by sMAPE, further demonstrates the accuracy of descriptor estimation.

**mmJoints on Specially-Crafted Models.** We further assess mmJoints's performance on pre-trained models with poor accuracy or nonfunctional models. We construct three models using the MARS architecture: $M_1$ for upper body pose, $M_2$ for lower body pose, and $M_3$ with random joints (randomly initialized, untrained parameters). Table 8 reveals $M_3$ generates high MPJPE and MAE across all joints, while $M_1$ and $M_2$ demonstrate lower MPJPE and MAE only for their respective target body parts. All models are trained (except $M_3$) and evaluated on the Self-Collected dataset.

After applying mmJoints to these three models, the estimated descriptors exhibit high accuracy for all three models. As shown in Table 9, the sMAPE and wMAPE for the sensing score $\xi$ remain low, indicating an accurate descriptor estimation, ranging from 2.48% to 2.74% for sMAPE and 3.42% to 3.48% for wMAPE. A similar trend is observed in reliability score. These results show that mmJoints operates effectively without requiring high accuracy from the pre-trained, black-box model, demonstrating robustness to variations in descriptor estimation.

**mmJoints Training and Inference Time.** The overhead of mmJoints's training and inference remains efficient even on a machine with limited computational resources. For example, on the Self-Collected dataset (containing 35,046 signal instances), feature modeling takes 7.5 hours, while the training time for signal distribution construction, surrogate pose construction, and descriptor estimation combined takes an additional 25 hours.

mmJoints has an average inference time–from signal input to estimating $(x, y, z, \xi, \kappa)$–of 23.12 ms. These measurements were performed on a system equipped with a GTX 1080 GPU and an i7-7800X CPU.

### Table 10: Ablation evaluation of component impact.

(a) Using $p_{refine}$ and signal feature improve descriptor estimation.

| Model Input | MSE ($\times 10^{-4}$) |
|---|---|
| $p_{pred}$ | 28.89 |
| $p_{pred}$ + signal | 9.09 |
| $p_{pred}$ + signal + $p_{refine}$ | 6.84 |

(b) Component analysis of how distribution improves the performance of surrogate pose construction.

| Configurations | Pre-trained | Pre-trained + Divergence | Pre-trained + Divergence + Signal Feature |
|---|---|---|---|
| MSE ($\times 10^{-2}$) | 9.66 | 4.41 | 2.67 |

### Table 11: mmJoints improves pose estimation performance across different settings.

(a) Evaluation in cross-environmental settings.

| Model | MPJPE (cm) before / after | MAE (cm) before / after | PCK@0.5 (%) before / after |
|---|---|---|---|
| MARS | 9.36 / **9.06** | 4.42 / **4.28** | 72.24 / **73.98** |
| mmBody | 9.23 / **8.89** | 4.38 / **4.22** | 72.92 / **75.78** |
| mmMesh | 8.07 / **7.83** | 3.81 / **3.70** | 79.29 / **80.99** |
| SynMotion | 11.69 / **11.21** | 5.65 / **5.38** | 55.81 / **59.79** |
| $M_1$ | 23.43 / **9.11** | 10.55 / **4.34** | 47.10 / **73.18** |
| $M_2$ | 41.71 / **8.84** | 17.86 / **4.19** | 27.49 / **75.33** |
| $M_3$ | 55.94 / **8.71** | 24.14 / **4.19** | 1.17 / **76.10** |

(b) Evaluation in cross-subject settings.

| Model | MPJPE (cm) before / after | MAE (cm) before / after | PCK@0.5 (%) before / after |
|---|---|---|---|
| MARS | 10.32 / **10.19** | 4.86 / **4.83** | 61.34 / **62.31** |
| mmBody | 9.59 / **9.33** | 4.59 / **4.48** | 64.03 / **65.88** |
| mmMesh | 8.59 / **8.50** | 4.15 / **4.08** | 70.68 / **70.97** |
| SynMotion | 11.06 / **10.75** | 5.19 / **5.03** | 55.12 / **57.25** |
| $M_1$ | 23.43 / **10.13** | 10.43 / **4.79** | 39.73 / **62.44** |
| $M_2$ | 40.42 / **9.64** | 17.15 / **4.58** | 24.22 / **65.21** |
| $M_3$ | 53.90 / **9.69** | 22.94 / **4.61** | 1.47 / **65.18** |

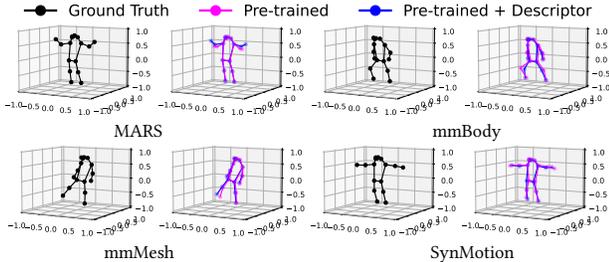

Figure 11: Examples of improved pose estimation from the self-collected dataset across four state-of-the-art baselines.

## 7.4 Ablation Study

**Ablation on Descriptor Estimation.** We explore the impact of the surrogate pose $p_{refine}$ and features of current signal frame in descriptor estimation by comparing with different baselines. We compare three baselines using the MARS model and its dataset: ① the vanilla baseline that uses only the predicted pose features from pre-trained model, $p_{pred}$, as input; ② a baseline that uses both $p_{pred}$ and signal features as input; and ③ the proposed approach, which uses $p_{pred}$, $p_{refine}$ and signal features as input. As shown in Table 10 (a), introducing the signal features improves descriptor estimation, reducing the error by 68.53% compared to the vanilla baseline ①. Further incorporating $p_{refine}$ reduces the error rate even more from baseline ②, as evidenced by an additional 24.76% reduction in MSE.

**Ablation on Surrogate Pose Construction.** We further analyze the impact of distribution and signal features in surrogate pose construction. As shown in Table 10 (b), directly using features from the pre-trained model's predicted pose as the surrogate pose results in a significantly larger error compared to the design that incorporates latent signal distribution information as input, which shows a 54.35% improvement in the MSE measurement. Furthermore, adding signal information leads to an additional 39.46% improvement.

## 8 APPLICATION EVALUATION

### 8.1 Application 1 – Improving Pose Estimation Accuracy

We apply mmJoints to pose estimation to improve the accuracy of the pre-trained model using the estimated descriptors. Descriptors are concatenated with the pre-trained model's output, adjusting only the input layer of the downstream model for fairness.

**Improved Pose Estimation.** We apply mmJoints to several pre-trained baselines, including four state-of-the-art models and three specially-crafted models (in Section 7.3), to evaluate its impact on improving pose estimation accuracy. Integrating descriptors into a lightweight MLP-based downstream model with pre-trained model's output improves pose estimation. Table 11 shows the results of all seven models on the Self-Collected dataset before and after applying mmJoints in cross-environment and cross-subject evaluations. Integrating the estimated descriptors generated by mmJoints consistently reduces joint localization error across all baselines and settings. mmJoints provides greater improvement (up to 6.42× improvement in MPJPE) when the model performs poorly, achieving results comparable to state-of-the-art baselines. Even for state-of-the-art pre-trained models, mmJoints improves the estimation accuracy across all evaluation settings.[3] A similar trend is observed in MAE and PCK@0.5.

We also present representative examples of pose instances involving upper and lower limb movements across four state-of-the-art baselines, as shown in Figure 11. After integrating descriptor information into this downstream task, we observe that the featured movements (e.g., arm movement in the MARS example and leg movement in the mmMesh example) are refined to more precise positions, resulting in more accurate pose estimation.

---

[3]While our improvements in estimation on state-of-the-art models might appear modest, the error is averaged over 17 joints. In practice, only a few joints contribute the majority of the error. For example, in the MARS model, we observed a 34.22 cm improvement in the estimation of the left wrist (with an average relative improvement of 12.52%), despite a general MPJPE improvement of only 0.31 cm. Similar gains reported in recent work [17, 54] demonstrate that such enhancements are considered significant.

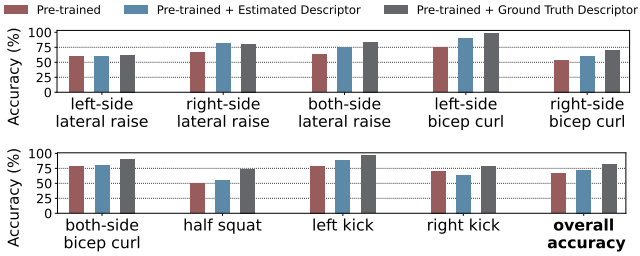

Figure 12: Accuracy comparison across various activities in cross-subject settings.

| Configurations | Pre-trained | Pre-trained + Estimated Descriptor | Pre-trained + Ground Truth Descriptor |
|---|---|---|---|
| Accuracy (%) | 55.83 | 61.00 | 72.86 |

(a) Overall accuracy comparison across different baselines.

| Joint Type | Correct | | Incorrect | |
|---|---|---|---|---|
| | Not Sensed | Sensed | Not Sensed | Sensed |
| #Joint | 13,852 | 6,564 | 3,860 | 1,904 |

(b) Test cases where estimated descriptors correct mispredictions.

Table 12: Evaluation in cross-environmental settings.

## 8.2 Application 2 – Improving Activity Recognition Accuracy

We evaluate the application of mmJoints for activity recognition in health rehabilitation. We employ an LSTM-based neural network as the baseline downstream classifier, with inputs consisting of poses estimated by the pre-trained MARS model from five consecutive frames.

**Breakdown of Activities.** We analyze accuracy improvements across activities using three models: a baseline classifier that relies solely on the pre-trained model's predictions, an enhanced classifier that uses both predictions and estimated descriptors, and another enhanced classifier that uses ground truth descriptors instead. Figure 12 shows the accuracy for nine activities and overall performance in cross-subject evaluation settings. Adding joint descriptors improves overall recognition accuracy by 6.49% and enhances activities by expanding the joint representation, with improvements of up to 16.07%. Joint descriptors particularly improve accuracy for high-variability activities (e.g., both-side lateral raise), where they capture additional patterns missed by the pre-trained baseline. For low-variability activities (e.g., half squat), the improvement is less noticeable as the predicted pose accuracy plays a more crucial role, with minor imperfections in the estimated descriptor having a limited effect. Note that our objective function prioritizes the overall accuracy, so while performance may slightly decrease in a few cases (e.g., some right kick instances, which constitute less than 1.5% of the entire test dataset), these instances are rare and have minimal impact on overall accuracy (0.71%).

| Configurations | Pre-trained | Pre-trained + $\xi$ | Pre-trained + $\xi + \kappa$ |
|---|---|---|---|
| Accuracy (%) | 66.24 | 69.87 | 72.73 |

Table 13: Component analysis in activity recognition.

**Cross-Environment Evaluation.** We further evaluate the accuracy improvements in unseen environmental settings with more complex indoor fixtures. As shown in Table 12 (a), the estimated descriptor improves activity recognition accuracy by 5.17%, while the ground truth descriptor achieves even higher accuracy of 72.86%, as expected. This result further demonstrates the effectiveness and generalizability of mmJoints in activity recognition.

We also investigate the distribution of joint types to understand the increase in accuracy. In Table 12 (b), we observe a considerable number of joints with limited signal support (not sensed) in both correctly and incorrectly predicted categories. In this evaluation, 67.85% of correctly predicted joints and 66.97% of incorrectly predicted joints are not sensed. The high proportion of unsensed joints in both categories highlights how the estimated descriptors identify latent patterns, further supporting our findings in Section 2.

**Impact of Descriptors.** We explore the role of descriptors in activity recognition. Table 13 shows accuracy improvements in cross-subject settings: 3.63% with the sensing score and an additional 2.86% with the reliability score, further enhancing the accuracy of the baseline classifier.

## 9 RELATED WORK

**mmWave-based Applications.** mmWave-based sensing applications, such as gesture and activity recognition [26, 34, 46, 49], human counting [31, 44], tracking [13, 22], detection [12, 20], biometric imaging [39, 56], pose estimation [4, 54, 57], and sensor fusion [19, 33], have advanced rapidly. We aim to characterize pre-trained pose estimators by expanding the representation of the output joint explicitly, enhancing system reliability and trustworthiness.

**mmWave-based Human Pose Estimation.** Recent work in mmWave-based human pose estimation has utilized various learning-based methods to generate human pose skeletons or meshes from radar signals [4, 10, 54, 59], focusing on learning the pattern of single [4] or multiple [10, 54, 59] signal frames. To explicitly characterize mmWave-based pose estimation models, we introduce descriptors that offer knowledge-level interpretability while also enhancing the precision of state-of-the-art methods and downstream tasks.

**Explainable Representation in Pose Estimation.** Several explainable AI methods are proposed for human pose estimation [9, 14, 32, 48, 55] using different input modalities. For vision-based tasks, existing methods [14, 55] highlight joints that preserve specific dependencies using RGB image as input. For mmWave-based tasks, methods either leverage implicit information (e.g., body parts) to improve pose estimation [9, 32] or generate explicit body-part explanations for better downstream tasks [48]. However, they lack explicit interpretability for arbitrary black-box models' outputs.

## 10 CONCLUSION

We propose mmJoints, a framework designed to characterize the output of pre-trained, black-box pose estimators to expand the joint representation beyond $(x, y, z)$ coordinates. mmJoints operates using only mmWave signals as input, without the need for understanding the intricate architecture of the pre-trained model or relying on high accuracy standards from the pre-trained model. Through extensive evaluations across different datasets, environments, and subject activities, we demonstrate the effectiveness and generalizability of mmJoints and its benefits in improving accuracy of downstream tasks.